\RequirePackage[svgnames]{xcolor}
\documentclass[11pt,letterpaper]{mystyle}

\definecolor{MorandiBlue}{HTML}{E6F6FF}
\definecolor{MorandiBlueLink}{HTML}{416B86}
\tcbset{titlebox/.append style={colback=MorandiBlue,colframe=MorandiBlue}}
\renewcommand{\titlefont}{\centering\color{black}\normalfont\fontsize{21}{23}\selectfont}
\renewcommand{\headingfont}{\color{black}\bfseries\fontsize{13}{14}\selectfont}

\newcommand{\gtechdata}[2][]{\par\noindent\textbf{#1:} #2\par}

\usepackage[comma,authoryear,compress]{natbib}
\let\cite\citep
\hypersetup{
  colorlinks=true,
  citecolor=YaleBlue,
  linkcolor=black,
  urlcolor=MorandiBlueLink,
  pdftitle={Learning to Look Beyond Screens: Transferable Visual Observation from GUI Grounding},
  pdfauthor={Ruilin Yao, Xinyu Qiu, Tianyu Zou, Yi Rong, Shengwu Xiong, Shuheng Shen, Changhua Meng}
}
\usepackage[all]{hypcap}
\usepackage{multirow}
\usepackage{wrapfig}
\usepackage{float}
\usepackage{fvextra}
\tcbuselibrary{breakable}
\usepackage{array,tabularx,placeins,needspace,etoolbox}
\newcolumntype{N}[1]{>{\centering\arraybackslash}p{#1}}
\newcolumntype{L}{>{\raggedright\arraybackslash}X}
\newcommand{\TableSetup}{\normalsize\setlength{\tabcolsep}{4pt}%
  \renewcommand{\arraystretch}{1.12}}
\newcommand{\TableNote}[1]{\par\vspace{4pt}%
  {\normalsize\raggedright #1\par}}

\AtBeginEnvironment{table}{\setlength{\belowcaptionskip}{\baselineskip}}

\title{Learning to Look Beyond Screens: Transferable Visual Observation from GUI Grounding}
\runningtitle{Learning to Look Beyond Screens}

\author[]{%
  \href{https://mr-bigworth.github.io/}{Ruilin Yao}\textsuperscript{1,2,*},
  \href{https://openreview.net/profile?id=~Xinyu_Qiu1}{Xinyu Qiu}\textsuperscript{3,*},
  \href{https://openreview.net/profile?id=~Tianyu_Zou2}{Tianyu Zou}\textsuperscript{1},
  \href{https://openreview.net/profile?id=~Yi_Rong3}{Yi Rong}\textsuperscript{1},
  \href{https://openreview.net/profile?id=~Shengwu_Xiong1}{Shengwu Xiong}\textsuperscript{1},\\ \vspace{-2pt}
  \href{https://openreview.net/profile?id=~Shuheng_Shen2}{Shuheng Shen}\textsuperscript{4,\textdagger},
  \href{https://openreview.net/profile?id=~Changhua_Meng1}{Changhua Meng}\textsuperscript{4}
  \\
  \vspace{4pt}
  {\normalfont\small\color{black}
    \textsuperscript{1}Wuhan University of Technology, 
    \textsuperscript{2}Institute of Automation, Chinese Academy of Sciences (CAS)\\ \vspace{-2pt}
    \textsuperscript{3}Zhejiang University\qquad
    \textsuperscript{4}Ant Group
  }
}

\begin{document}
% This class collects the abstract before rendering the title panel.
\begin{abstract}
With the rise of agentic reasoning, visual perception is evolving from passive recognition toward active, task-oriented observation. Active visual perception requires an agent to decide not only what to recognize, but also where to look, how much context to inspect, and when further observation is worthwhile. We explore GUI grounding as a scalable source of spatially grounded supervision for learning such observation behaviors, with the goal of transferring them beyond interface-centric tasks. To this end, we introduce LBS (\textbf{L}ook \textbf{B}eyond \textbf{S}creens), a vision-language model for transferable active visual perception. Our framework first bootstraps observation capabilities from teacher-synthesized agentic GUI grounding trajectories and then optimizes multi-turn crop-and-inspect decisions with reinforcement learning. Spatial target annotations are transformed into outcome-gated rewards with selective credit assignment, allowing the model to learn not only how to inspect, but also which inspections are useful for task completion. Under fixed inference settings, models based on Qwen3.5-35B-A3B and Qwen3.8-27B obtain consistent improvements over their corresponding baselines. Crucially, after agentic reinforcement learning, these gains transfer to out-of-domain non-GUI benchmarks such as PerceptionBench, HR-Bench and ChartQAPro, exceeding those obtained from cold-start supervised fine-tuning alone. This result suggests that the learned capability is not merely GUI-specific imitation, but a more general observation strategy. Further analysis reveals a complementary role of the two training stages: SFT acquires the primitives of visual inspection, whereas RL learns to allocate observation effort according to downstream utility. We therefore establish GUI grounding as a scalable supervision paradigm for active visual perception and provide evidence that interface-centric interaction data can induce transferable observation behaviors for broader multimodal reasoning.
Our code and model are publicly available at \url{https://github.com/inclusionAI/UI-Venus/tree/LBS} and \url{https://huggingface.co/inclusionAI/LBS-27B}, respectively.
\end{abstract}

\maketitle
\begingroup
  \renewcommand{\thefootnote}{}% 去掉脚注编号
  \footnotetext{\textsuperscript{*}Core contributors.\quad\textsuperscript{\textdagger}Tech lead.}
\endgroup
\vspace{3mm}
\section{Introduction}

\gtechdata[Code]{\url{https://github.com/inclusionAI/UI-Venus/tree/LBS}}
\gtechdata[Model]{\url{https://huggingface.co/inclusionAI/LBS-27B}}

Active visual perception requires a vision-language model (VLM) to acquire evidence as well as interpret it. On a high-resolution image, the model must locate relevant regions, select an appropriate view, and relate local detail to the original scene~\cite{khayatkhoei2025mllms,bai2025qwen3}. Crop-and-inspect and think-with-image methods demonstrate the value of additional visual observations~\cite{qi2025cogcom,zheng2026deepeyes,wu2024vstar,wang2026traceable}. A central training challenge is to connect these intermediate observation decisions to task success, particularly when a correct final answer can follow either informative or redundant inspection.

GUI grounding provides a concrete setting for learning this connection. Each example pairs an instruction with a spatial target, making both the final localization and intermediate crop geometry measurable~\cite{cheng2024seeclick,xu2024aguvis,li2025screenspot}. These annotations can therefore supervise intermediate observation decisions without specifying an ideal crop sequence for reinforcement learning. We hypothesize that such supervision can help elicit latent visual knowledge in pretrained VLMs by training them to put this knowledge into practice through sequential observation: deciding where to inspect, how much context to retain, and when further evidence is useful. Although the source images are interfaces, these decisions are also relevant to natural images and charts. This motivates our central question: \emph{can spatial supervision from GUI grounding elicit observation behavior that transfers to non-GUI perception?} Such transfer would suggest a practical way to make pretrained visual knowledge more accessible through interaction, potentially reducing the need to construct specialized observation trajectories for each downstream domain.

We address this question with a two-stage framework. Supervised fine-tuning (SFT) on teacher-generated grounding trajectories initializes the crop-and-inspect protocol. Multi-turn reinforcement learning (RL) then refines the policy using grounding correctness and training-only target annotations~\cite{guo2025deepseek}. Outcome-gated spatial rewards connect intermediate inspections to task success, a small-target bonus encourages target coverage, and selective response masking controls which intermediate responses receive policy updates. In this way, spatial annotations provide supervision for the process of acquiring visual evidence as well as for the final grounding result. The same crop interface is used at inference across domains, with task-specific final answers. This design allows us to examine whether observation behavior learned under GUI supervision benefits tasks with different visual content and objectives.

Our results support transfer beyond supervised initialization. RL improves ScreenSpot-Pro \cite{li2025screenspot} accuracy over SFT by 1.83 and 6.13 percentage points for Qwen3.5-35B-A3B and Qwen3.8-27B, and improves PerceptionBench \cite{lin2026perceptionbench} by approximately 3.0 and 7.10 points. The Qwen3.5-based policy also improves on HR-Bench 4K/8K \cite{wang2024divide} and ChartQAPro \cite{masry2025chartqapro}. The reward comparison reveals a useful distinction: outcome-only RL improves source-domain grounding but lowers PerceptionBench accuracy, whereas the spatially supervised configuration improves both. These comparisons show that higher GUI grounding accuracy does not necessarily imply stronger non-GUI transfer; the results highlight the value of supervising intermediate observation decisions. Comparisons of reconstructed interaction histories further examine how crop images and their presentation scale affect answer generation. Together, these findings support GUI grounding as a practical source of supervision for visual observation whose benefits extend beyond the source domain, consistent with the hypothesis that spatially grounded interaction training can help elicit and apply pretrained visual knowledge across tasks.

Our contributions are threefold:
\begin{itemize}
\item We establish empirical evidence that GUI grounding can supervise visual observation policies that transfer beyond interfaces, with gains over SFT across two backbones on PerceptionBench and across additional high-resolution and chart benchmarks for Qwen3.5-35B-A3B.
\item We develop a two-stage framework that converts target annotations into outcome-gated spatial rewards and selective response updates for multi-turn crop-and-inspect training.
\item We show that improved grounding under outcome-only RL need not translate into improved non-GUI perception, and complement the training comparison with controlled edits to historical crop observations.
\end{itemize}

\section{Related Work}

\paragraph{Agentic visual reasoning.}
Agentic visual reasoning treats observation as sequential evidence acquisition. ReAct interleaves reasoning with environment interaction \cite{yao2023react}, and Toolformer learns API use from self-supervised data \cite{schick2023toolformer}. Visual agents similarly use image manipulation and code-based tools during reasoning \cite{team2026kimi,anthropic2026claudeopus48,google2026agenticvision}. CogCoM integrates image manipulations into reasoning \cite{qi2025cogcom}; V$^*$ uses guided visual search to select informative regions at inference time \cite{wu2024vstar}. DeepEyes and DeepEyesV2 study reinforcement learning for visual and multimodal tool use \cite{zheng2026deepeyes,hong2026deepeyesv2}. These methods establish the utility of additional observations and tool interaction. Our focus is the supervision that teaches inspection behavior and whether it transfers from GUI grounding to non-GUI perception.

\paragraph{Visually grounded reasoning.}
Visually grounded reasoning connects linguistic predictions to image regions~\cite{kazemzadeh2014referitgame,plummer2015flickr30k,zhu2016visual7w}.
Recent methods can be organized along two axes: by paradigm, into training-based approaches~\cite{sarch2026grounded,zheng2026deepeyes,wang2026vgr} and test-time approaches~\cite{khayatkhoei2025mllms,shen2025zoomeye,zhang2026cofft}; and by motivation, into suppressing irrelevant background~\cite{wei2026zooming,liu2025hide} and enhancing task-relevant foreground through zoom-in operations~\cite{fan2026grit,luo2025visual,yuan2026vision,zhang2026finers,zheng2026deepeyes}.
ViGoRL anchors reasoning to visual coordinates, while TreeVGR jointly supervises localization and reasoning \cite{sarch2026grounded,wang2026traceable}. GUI grounding methods instead emphasize instruction-conditioned target localization and downstream action execution \cite{cheng2024seeclick,xu2024aguvis,venusteam2026uivenus2technicalreport}. Our study connects these directions by using GUI target annotations to supervise intermediate observation actions and evaluating the resulting policy on non-GUI perception. The central distinction is the source of supervision and its transfer: one spatially annotated interaction domain supplies feedback for observation policies evaluated across different visual content and answer formats.

\section{Method}
\label{sec:method}

\begin{figure}[t]
  \centering
  \includegraphics[width=\linewidth]{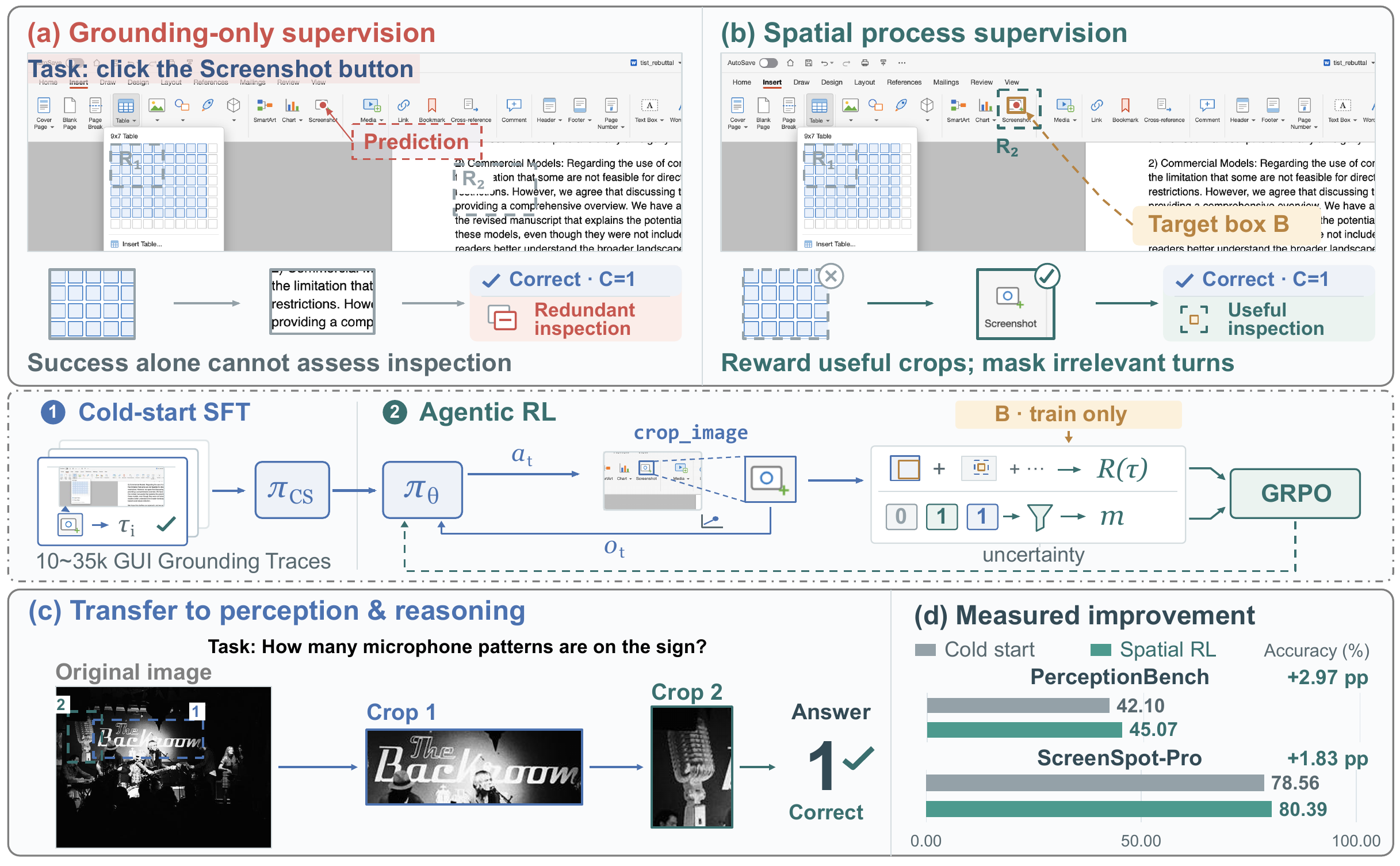} \vspace{-7pt}
  \caption{Learning transferable observation from GUI grounding. (a) Grounding success alone cannot distinguish target-relevant from redundant inspection. (b) Following SFT, spatially supervised RL uses training-only target annotations to score crops and mask irrelevant intermediate responses. (c) The observation policy applies the same crop-and-inspect protocol to non-GUI images. (d) The Qwen3.5-based model improves over SFT by 2.97 and 1.83 percentage points on PerceptionBench and ScreenSpot-Pro, respectively.}
  \label{fig:intro1}\vspace{-5pt}
\end{figure}

Our framework turns GUI target annotations into supervision for sequential visual observation (Figure~\ref{fig:intro1}). SFT first teaches the interaction protocol; RL then scores crop geometry and filters intermediate response updates using grounding outcomes. The policy receives images and instructions, while target annotations remain training-only information. At inference, the crop interface is shared across domains and the final answer format is adapted to each task.

\subsection{Grounding Supervision and Policy Initialization}
\label{sec:cold_start}

\paragraph{Synthetic grounding tasks.}
We construct interfaces through code-guided generation, following prior work on synthetic multimodal and GUI data~\cite{yang2025scaling,venusteam2026uivenus2technicalreport}. An LLM produces executable HTML/CSS/JavaScript from a scenario, optionally conditioned on a persona or reference screenshot. Rendering associates each screenshot with DOM structure, element states, and geometry. Candidate targets are extracted from element semantics and accessibility attributes, filtered for visibility, clipping, occlusion, and correspondence to rendered content, and refined using text-node geometry where appropriate. Accepted targets are paired with natural-language interaction instructions; infeasible instructions provide negative examples. Both SFT and RL use only GUI grounding instructions, with no general-domain VQA training tasks. Reference screenshots used for synthetic image generation exclude images from the evaluation benchmarks.

A grounding task is $d=(I,q,B)$, where $I$ is the screenshot, $q$ is the instruction, and $B=(b_x^-,b_y^-,b_x^+,b_y^+)$ is the target box in the original image's $[0,1000]^2$ coordinate system. Infeasible tasks use $B=(-1,-1,-1,-1)$. The policy observes only $(I,q)$; $B$ is reserved for trajectory filtering and training feedback.

\paragraph{Teacher trajectories.}
We use Qwen3.8-Max~\cite{qwen2026qwen38} to generate multi-turn trajectories at temperature $1.0$, with a maximum generation length of $32{,}768$ tokens. Each trajectory $\tau=(a_1,o_1,\ldots,a_{T-1},o_{T-1},a_T)$ contains assistant responses $a_t$ and tool observations $o_t$. The grounding response predicts a point $\hat p=(\hat x,\hat y)$. For a parseable finite prediction, correctness is
\begin{equation}
C(d,\tau)=\mathbf{1}\{b_x^-\leq\hat x\leq b_x^+,\ b_y^-\leq\hat y\leq b_y^+\}.
\label{eq:correctness}
\end{equation}
Invalid predictions receive zero. The same rule accepts only $(-1,-1)$ for an infeasible task. Although the prompt requests an element center, acceptance requires point-in-box agreement, including the boundary. Rejection sampling retains successful trajectories:
\begin{equation}
\mathcal D_{\mathrm{CS}}=\{(d,\tau):\tau\sim\pi_{\mathrm{teach}}(\cdot\mid I,q),\ C(d,\tau)=1\}.
\label{eq:cold_data}
\end{equation}
Rejection sampling ensures grounding correctness but leaves intermediate inspection quality unconstrained, motivating the spatial feedback used during RL. 

\paragraph{Supervised initialization.}
We train the student with Swift~\cite{zhao2025swift}, supervising assistant outputs while conditioning on preceding images, responses, and tool feedback. With $\mathcal Y(\tau)$ denoting assistant-token positions, the objective is
\begin{equation}
\mathcal L_{\mathrm{CS}}(\theta)\propto-\sum_{(d,\tau)\in\mathcal D_{\mathrm{CS}}}\sum_{k\in\mathcal Y(\tau)}\log\pi_\theta(y_k\mid I,q,\tau_{<k}).
\label{eq:sft}
\end{equation}
The resulting checkpoint initializes RL and provides an inspection-capable baseline, allowing us to evaluate improvement beyond imitation of the tool-use protocol.

\subsection{Multi-Turn Visual Observation}
\label{sec:agentic_rl}

At turn $t$, the policy samples $a_t\sim\pi_\theta(\cdot\mid h_t)$, where $h_t$ contains the task, original image, and interaction history. Each response contains reasoning followed by either a final answer or one or two textual \texttt{crop\_image} actions. A crop action specifies an available image and a rectangle in its local pixel coordinates; previously returned crops may be inspected recursively. Final grounding predictions always use original-image normalized coordinates.

The tool returns the crop image, its dimensions, and a coordinate mapping to the original image. Results from the same turn are added to the history in action order; failed executions return textual feedback. Earlier images and messages remain available, so each crop adds a local view while preserving access to the original context. Interaction ends with a final answer or a configured stopping limit. Every admitted crop attempt consumes a tool-call slot, including failed executions. Appendix~\ref{app:implementation} details crop construction and coordinate tracking.

\subsection{Outcome-Gated Spatial Rewards}

Grounding correctness assigns the same outcome to successful trajectories with different inspection quality. We use target annotations in two complementary ways: a spatial reward evaluates the crop sequence, and the response mask in Section~\ref{sec:selective_updates} controls which intermediate outputs receive direct updates. Let $R_1,\ldots,R_K$ be the successful crop regions in execution order, mapped to original-image coordinates, and let $u_j=\operatorname{IoU}(R_j,B)$.

\paragraph{Spatial process reward.}
We reward the best target overlap and penalize repeated crops, declining overlap, and tool use:
\begin{align}
P_{\mathrm{rep}}&=\min\left(0.4,\;0.2\sum_{j=2}^{K}\mathbf{1}\{\operatorname{IoU}(R_{j-1},R_j)>0.9\}\right),\\
P_{\mathrm{reg}}&=0.2\,\mathbf{1}\{K\geq2\ \text{and}\ u_K<u_1\},\\
R_{\mathrm{proc}}&=C(d,\tau)\,\operatorname{clip}_{[0,1]}\left(\max_j u_j-P_{\mathrm{rep}}-P_{\mathrm{reg}}-0.05K\right).
\label{eq:process}
\end{align}
The reward is zero for infeasible tasks or trajectories without successful crops. Repetition compares consecutive crops, whereas regression compares the final and initial target overlaps. The outcome gate restricts spatial rewards to successful trajectories, and clipping bounds this component between zero and one. Crop overlap supplies a proxy for spatial relevance; it does not directly evaluate interpretation of the returned image.

\paragraph{Small-target inspection bonus.}
IoU can be low when a useful contextual crop contains a small target. We therefore define a turn-level relevance score that combines overlap and target-center containment:
\begin{equation}
Q_t=\max_{j:\,R_j\text{ produced at }t}\left[0.6\operatorname{IoU}(R_j,B)+0.4\mathbf{1}\{c_B\in R_j\}\right],
\label{eq:quality}
\end{equation}
where $c_B$ is the target center. A crop containing this center receives at least $0.4$, even when its overlap is small. For targets occupying less than $5\%$ of the image, we add
\begin{equation}
R_{\mathrm{zoom}}=0.2\,C(d,\tau)\,\mathbf{1}\{|B|/10^6<0.05\}\,\mathbf{1}\{\max_t Q_t\geq0.2\}.
\label{eq:zoom}
\end{equation}
This bonus is zero for infeasible tasks and trajectories without successful crops. It is a target-coverage incentive: a broad crop containing the target center can qualify, while crop selectivity is encouraged separately by the overlap-based process reward. The bonus depends on geometry rather than the image-resizing factor.

\paragraph{Format and textual rewards.}
The format reward $R_{\mathrm{fmt}}$ scores the final response: canonical parsing receives $1$, compatibility parsing receives $0.5$, and parsing failure receives $0$. We additionally examine a text-based critic that evaluates task focus, feedback use, reasoning--action consistency, incremental progress, and conciseness. Given the task and a textual trajectory, the critic returns $s_{\mathrm{cot}}\in[0,1]$, yielding
\begin{equation}
R_{\mathrm{cot}}=C(d,\tau)\,s_{\mathrm{cot}}.
\label{eq:critic}
\end{equation}
Trajectories without visible reasoning receive zero. The non-oracle critic receives neither images nor target annotations, so its score measures textual consistency rather than visual correctness. The general reward is
\begin{equation}
R(d,\tau)=C(d,\tau)+0.3R_{\mathrm{proc}}+0.1R_{\mathrm{fmt}}+R_{\mathrm{zoom}}+0.3R_{\mathrm{cot}}.
\label{eq:reward}
\end{equation}
All components contribute to one trajectory-level return.

\subsection{Selective Response Updates and Optimization}
\label{sec:selective_updates}

\paragraph{Filtering intermediate responses.}
A shared trajectory reward can reinforce irrelevant actions in an otherwise successful rollout. We use the turn-level relevance score to construct a response mask:
\begin{equation}
m_t=\begin{cases}
0,&t<T,\ C(d,\tau)=1,\ Q_t\text{ exists},\ Q_t<0.05,\\
1,&\text{otherwise}.
\end{cases}
\label{eq:mask}
\end{equation}
Equation~\eqref{eq:mask} defines the spatial mask. A second filter removes spatially eligible intermediate responses whose mean generated-token surprisal is below the median over eligible turns, using scaling factor $\eta=1$. Let $e_t\in\{0,1\}$ denote this second filter and $\widetilde m_t=m_t e_t$ the resulting response mask, with $\widetilde m_T=1$. Both filters act on direct response loss; all previous responses and observations remain in the conditioning history. Filtering is independent of the advantage sign; it selects direct response updates without assigning a separate causal value to each crop. Appendix~\ref{app:implementation} specifies the filtering procedure.

\paragraph{Group-relative optimization.}
We adapt GRPO~\cite{shao2024deepseekmath} to multi-turn trajectories. For each task, $G=8$ rollouts receive the reward in Equation~\eqref{eq:reward}. Each advantage subtracts the group mean without variance normalization and is shared across retained assistant tokens. Training uses a PPO-style clipped surrogate~\cite{schulman2017proximal} with a fixed-reference KL penalty, averaging first over retained tokens and then over trajectories.

\section{Experiments}
\label{sec:experiments}
\subsection{Experimental Setup}
\label{sec:exp_setup}
\paragraph{Training configuration.}
We study Qwen3.5-35B-A3B~\cite{qwen2026qwen35} and Qwen3.8-27B~\cite{qwen2026qwen38}. For both models, we freeze the vision encoder and vision--language projector throughout SFT and RL, updating only the language-model parameters. Qwen3.5-35B-A3B initially exhibits limited adherence to our prompting format and agentic observation protocol. To establish reliable protocol following, we use approximately $35{,}000$ SFT trajectories for Qwen3.5-35B-A3B, and $10{,}000$ for Qwen3.8-27B. Each SFT checkpoint then initializes the corresponding RL policy. The main configuration combines grounding correctness, gated spatial feedback, format compliance, small-target inspection bonus and the auxiliary critic. Optimization settings and prompts appear in Appendices~\ref{app:implementation} and~\ref{app:prompts}. We also keep the training recipe simple and do not extensively tune the remaining optimization hyperparameters, focusing our study on the proposed supervision design.

\paragraph{Interaction protocol.}
The crop tool uses a scale factor of $\alpha=4.0$ and a pixel cap of $P_{\max}=4\times10^6$ (Equation~\eqref{eq:resize}). Original images are preprocessed with a pixel count between $65{,}536$ and $16{,}777{,}216$. We generate rollouts using vLLM~\cite{kwon2023efficient}. Training allows up to three crop calls per example. Qwen3.8-27B evaluation also uses at most three crop calls, with \texttt{reasoning\_effort=xhigh} and \texttt{thinking\_budget=4096} ($4{,}096$ reasoning tokens per assistant turn), passed through \texttt{chat\_template\_kwargs}. Within each benchmark, SFT and RL models use identical prompts, decoding settings, and inference limits.

\paragraph{Benchmarks and metrics.}
ScreenSpot-Pro \cite{li2025screenspot} evaluates instruction-conditioned GUI localization using point-in-box accuracy. PerceptionBench \cite{lin2026perceptionbench} evaluates non-GUI perception across ten capabilities using reference-based answer judgment, with a locally deployed DeepSeek-V4-Flash \cite{xu2026deepseek} serving as the judge model. The Qwen3.8-27B evaluation uses Deepseekv4-flash-0731, with no failed or skipped judgments across all $3{,}000$ records. We interpret stage-wise gains within each backbone rather than treating cross-backbone scores as a controlled ranking. We report overall accuracy and category-level scores. Additional evaluations cover OSWorld-G-R (Refine) \cite{xie2025scaling}, HR-Bench \cite{wang2024divide} at 4K and 8K resolutions, and ChartQAPro~\cite{masry2025chartqapro}. 
Further details on selected additional benchmarks are provided in Appendix~\ref{app:breakdown}.
\begin{table}[t]
\centering
\caption{ScreenSpot-Pro accuracy (\%) by application group.}\vspace{-5pt}
\label{tab:sspro_results}
\TableSetup
\small
\renewcommand{\arraystretch}{0.95}
\setlength{\tabcolsep}{3pt}
\begin{tabularx}{\linewidth}{@{}L c *{7}{c}@{}}
\toprule
& & \multicolumn{6}{c}{Application group} & \\
\cmidrule(lr){3-8}
Model & Zoom & Dev. & Creat. & CAD & Sci. & Office & OS & Overall \\
\midrule
Holo2-235B-A22B (Agentic)
& \ding{55} & 74.6 & 72.1 & 80.5 & 79.9 & 85.2 & 83.2 & 78.5 \\
MAI-UI-32B (MVP)
& \ding{51} & 79.3 & 69.5 & 76.6 & 76.0 & 91.3 & 75.5 & 77.5 \\
AdaZoom (Qwen3.5-397B refine)
& \ding{51} & 75.3 & 72.4 & 72.4 & 74.0 & 93.5 & 76.5 & 76.8 \\
KV-Ground-GuiOwl-1.5 (Zoom In)
& \ding{51} & 73.6 & 72.4 & 74.7 & 77.2 & 90.0 & 73.0 & 76.4 \\
MAI-UI-32B (Zoom In)
& \ding{51} & 71.6 & 70.4 & 72.8 & 75.6 & 86.1 & 65.3 & 73.5 \\
\midrule
Base (Qwen3.5-35B-A3B)
& \ding{51} & -- & -- & -- & -- & -- & -- & $1.27^{*}$ \\
Ours (+SFT, 35k trajectories)
& \ding{51} & 77.26 & 70.67 & 78.93 & 81.10
& \textbf{90.00} & 77.04 & 78.56 \\
Ours (+Agentic RL)
& \ding{51} & \textbf{83.61} & \textbf{71.55}
& \textbf{80.08} & \textbf{83.07} & 87.83
& \textbf{79.08} & \textbf{80.39} \\
\midrule
Base (Qwen3.8-27B)
& \ding{51} & 69.23 & 61.88 & 67.05 & 67.32 & 83.48 & 70.92 & 69.26 \\
Ours (+SFT, 10k trajectories)
& \ding{51} & 76.92 & 68.33 & 70.11 & 74.41 & 86.96 & 77.04 & 75.02 \\
Ours (+Agentic RL)
& \ding{51} & \textbf{82.61} & \textbf{74.78}
& \textbf{77.39} & \textbf{80.71} & \textbf{93.04}
& \textbf{81.63} & \textbf{81.15} \\
\bottomrule
\end{tabularx}
\TableNote{Zoom indicates zoom-in use. Bold denotes the highest reported accuracy
within each backbone and column; dashes indicate unreported scores.
$^{*}$ The base Qwen3.5-35B-A3B model does not reliably follow our
prompting format and interaction protocol.
}\vspace{-12pt}
\end{table}
\subsection{GUI Grounding Across Backbones}
\label{sec:main_results}
We compared our method with several mainstream GUI grounding methods \cite{hai2025holo2modelfamily,zhang2026mvp,kim2026zoom,pei2026adazoom,zhou2025mai}, and the results are presented in Table ~\ref{tab:sspro_results}. Table~\ref{tab:sspro_results} shows that RL improves ScreenSpot-Pro accuracy over SFT for both backbones. Qwen3.5-35B-A3B increases from $78.56\%$ to $80.39\%$, and Qwen3.8-27B increases from $75.02\%$ to $81.15\%$. Since both SFT baselines already follow the crop-and-inspect protocol, these gains support the value of optimizing observation trajectories beyond teacher imitation.
Qwen3.8-27B gains $11.89$ points over base and improves all six application groups over SFT, led by CAD and engineering ($+7.28$), creative applications ($+6.45$), and scientific applications ($+6.30$). Qwen3.5-35B-A3B improves in five groups, led by development ($+6.35$), but declines in office applications ($90.00\%$ to $87.83\%$). These results are consistent with the utility of zoom-in inspection for high-resolution GUI grounding, where enlarged local views can make small interface elements and fine-grained visual details easier to distinguish and localize.

\subsection{Transfer to General Visual Perception}
\begin{table}[!htbp]\vspace{-10pt}
\centering 
\caption{PerceptionBench accuracy (\%) across training stages
for two backbones.}
\label{tab:pb_results_combined}
\TableSetup
\renewcommand{\arraystretch}{0.9}
\small
\setlength{\tabcolsep}{3pt}
\begin{tabularx}{\linewidth}{
    @{}
    >{\raggedright\arraybackslash}p{0.25\linewidth}
    *{4}{>{\centering\arraybackslash}X}
    |
    *{4}{>{\centering\arraybackslash}X}
    @{}
}
\toprule
& \multicolumn{4}{c|}{Qwen3.5-35B-A3B}
& \multicolumn{4}{c}{Qwen3.8-27B} \\
\cmidrule(lr){2-5}
\cmidrule(lr){6-9}
Capability
& Base & +SFT & +RL & \shortstack{Gain}
& Base & +SFT & +RL & \shortstack{Gain} \\
\midrule
Visual relation
& 40.0 & 39.7 & \textbf{45.2} & +5.5
& 53.9 & 59.70 & \textbf{68.48} & +8.78 \\
Counting
& 41.3 & 42.5 & \textbf{48.8} & +6.3
& 48.5 & 53.94 & \textbf{61.82} & +7.88 \\
Attribute
& 40.3 & 47.3 & \textbf{48.5} & +1.2
& 50.6 & 55.45 & \textbf{65.76} & +10.31 \\
Depth / 3D perception
& 33.1 & \textbf{37.6} & \textbf{37.6} & 0.0
& 39.7 & \textbf{50.91} & 50.30 & -0.61 \\
Localization
& 37.0 & \textbf{51.3} & 50.0 & -1.3
& 48.5 & 65.45 & \textbf{67.58} & +2.13 \\
Comparison
& 39.8 & 35.5 & \textbf{43.1} & +7.6
& 49.5 & 51.61 & \textbf{63.44} & +11.83 \\
Fine-grained recognition
& 41.4 & 41.8 & \textbf{47.0} & +5.2
& 43.4 & 52.07 & \textbf{57.59} & +5.52 \\
Contextual integration
& 34.2 & \textbf{38.5} & 38.1 & -0.4
& 38.8 & 44.31 & \textbf{52.55} & +8.24 \\
OCR
& 53.0 & 58.5 & \textbf{60.4} & +1.9
& 50.6 & 53.73 & \textbf{61.57} & +7.84 \\
Hallucination
& 26.2 & 28.1 & \textbf{31.8} & +3.7
& 41.0 & 33.21 & \textbf{43.91} & +10.70 \\
\midrule
\textbf{Overall}
& 38.6 & 42.1 & \textbf{45.1} & +3.0
& 46.6 & 52.57 & \textbf{59.67} & +7.10 \\
\bottomrule
\end{tabularx}
\TableNote{Gain is the accuracy difference between +RL and +SFT in percentage points.}
\vspace{-5pt}
\end{table}

Table~\ref{tab:pb_results_combined} shows improvements beyond SFT for both backbones on PerceptionBench. Qwen3.5-35B-A3B increases from $38.6\%$ at base to $42.1\%$ after SFT and $45.1\%$ after RL; Qwen3.8-27B follows the same progression, from $46.6\%$ to $52.57\%$ and $59.67\%$. The RL gains of approximately $3.0$ and $7.10$ points show that non-GUI improvement is not confined to one backbone. Both final policies exceed their respective base models across all ten capabilities.

The capability profiles reveal shared benefits and model-dependent tradeoffs. Relative to SFT, both policies improve counting, comparison, fine-grained recognition, OCR, and hallucination assessment. For Qwen3.5-35B-A3B, the largest gains are in comparison ($+7.6$), counting ($+6.3$), and visual relation ($+5.5$); RL also reverses the SFT declines in comparison and visual relation. Qwen3.8-27B gains most in comparison ($+11.83$), hallucination assessment ($+10.70$), and attribute recognition ($+10.31$). Qwen3.5-35B-A3B improves seven of ten capabilities, while Qwen3.8-27B improves nine, with a slight decline in depth/3D perception ($50.91\%$ to $50.30\%$, $-0.61$ points). 
The shared gains extend beyond grounding to tasks requiring discrimination and comparison of visual evidence. These results support transfer of the trained observation policy while showing that its benefits depend on the perceptual capability. Appendix~\ref{app:pb_external} provides external reference scores.

\subsection{Transfer Across Visual Domains}
\label{sec:additional_context}

\begin{figure}[t]
  \centering
  \includegraphics[width=0.75\linewidth]{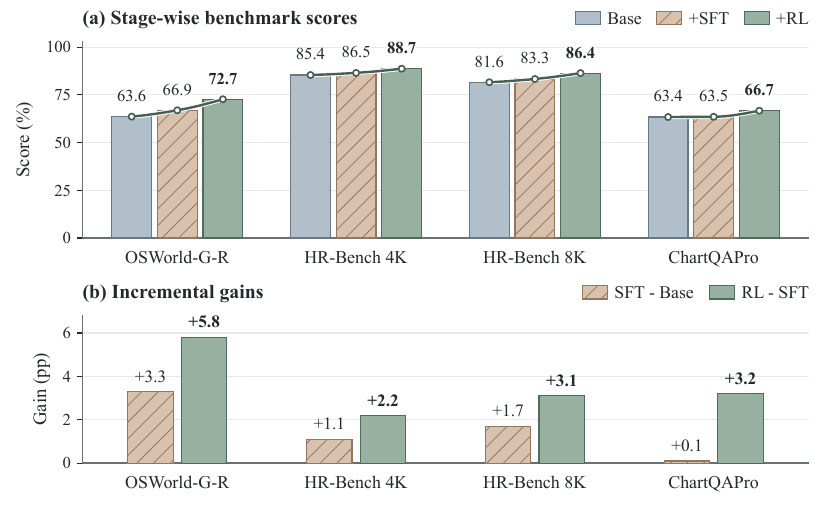}\vspace{-8pt}
  \caption{Stage-wise results for Qwen3.5-35B-A3B.}\vspace{-8pt}
  \label{fig:stagewise_transfer_curves}
\end{figure}
Figure~\ref{fig:stagewise_transfer_curves}(a) extends the stage-wise
evaluation of Qwen3.5-35B-A3B to another GUI setting, high-resolution
perception, and chart understanding. Scores increase from Base to
SFT to RL on all four benchmarks. On OSWorld-G-R, SFT raises accuracy
from $63.6\%$ to $66.9\%$, and RL further improves it to $72.7\%$,
yielding a $5.8$-point gain over SFT.
Figure~\ref{fig:stagewise_transfer_curves}(b) compares the incremental
gains from the two training stages. RL yields larger gains than SFT
on every benchmark. On the non-GUI tasks, RL improves over SFT by
$2.2$, $3.1$, and $3.2$ percentage points on HR-Bench 4K, HR-Bench 8K,
and ChartQAPro, respectively, exceeding the corresponding SFT gains
over Base of $1.1$, $1.7$, and $0.1$ points. ChartQAPro illustrates
this difference: SFT changes the score only from $63.4\%$ to
$63.5\%$, whereas RL raises it to $66.7\%$. Together with
PerceptionBench, these results show that improvements beyond
supervised initialization extend across general perception,
high-resolution imagery, and charts under the reported evaluation
settings.

\subsection{Reward Design Ablation}
\label{sec:ablation}
\begin{table}[!htbp]
\centering\vspace{-8pt}
\caption{Source-domain and transfer accuracy (\%) for Qwen3.5-35B-A3B reward configurations.}\vspace{-2pt}
\label{tab:ablation}
\TableSetup
\begin{tabularx}{\linewidth}{@{}L c c c *{2}{N{48pt}}@{}}
\toprule
& \multicolumn{3}{c}{Reward components} & \multicolumn{2}{c}{Accuracy (\%)} \\
\cmidrule(lr){2-4}\cmidrule(lr){5-6}
Configuration & Process & Bonus & Critic & {SSPro} & {PB} \\
\midrule
Outcome-only RL & \ding{55} & \ding{55} & \ding{55} & 79.19 & 40.87 \\
Ungated process + critic & Ungated & \ding{55} & \ding{51} & \bfseries 81.09 & 42.83 \\
Ungated process + bonus & Ungated & \ding{51} & \ding{55} & 79.57 & 42.10 \\
Gated process & Gated & \ding{55} & \ding{55} & 79.25 & 41.63 \\
Gated process + bonus  & Gated & \ding{51} & \ding{55} & 79.70 & 43.26 \\
Gated process + critic & Gated & \ding{55} & \ding{51} & 79.82 & 42.57 \\
\midrule
\textbf{Gated process + bonus + critic (ours)} & Gated & \ding{51} & \ding{51} & 80.39 & \bfseries 45.07 \\
\bottomrule
\end{tabularx}
\TableNote{Outcome-only RL uses only correctness and format rewards. Process: spatial process reward; Bonus: small-target inspection bonus; Critic: text-based CoT critic. }
\end{table}

Outcome-only RL reveals a distinction between source-domain improvement and cross-domain transfer. It reaches $79.19\%$ on
ScreenSpot-Pro and $40.87\%$ on PerceptionBench, compared with $78.56\%$ and $42.10\%$ for SFT. Thus, a $0.63$-point grounding
gain accompanies a $1.23$-point transfer decline. Our full configuration, combining the gated spatial process reward,
small-target inspection bonus, and text-based critic, improves
both benchmarks over SFT and exceeds outcome-only RL by
$1.20$ and $4.20$ points, respectively. The configuration with the highest ScreenSpot-Pro accuracy is not the one with the strongest transfer. These results indicate
that better source-task performance need not translate into
better non-GUI perception, motivating supervision of intermediate
observation behavior beyond grounding correctness.

The component comparisons highlight the role of the combined
reward design (Table~\ref{tab:ablation}). Our full configuration
achieves the highest PerceptionBench accuracy ($45.07\%$),
whereas ungated process reward with the critic achieves the
highest ScreenSpot-Pro accuracy ($81.09\%$).
Removing the critic from the full configuration lowers
ScreenSpot-Pro/PerceptionBench accuracy by $0.69$/$1.81$ points;
removing the bonus lowers accuracy by $0.57$/$2.50$ points.
These comparisons support retaining both components, with
larger observed gains on the transfer benchmark.
The effect of outcome gating depends on the accompanying
components: gating improves both benchmark scores when paired
with the bonus, but lowers both when paired with the critic
alone. Taken together, the results favor the full configuration
for cross-domain transfer, while showing that gating alone
does not consistently improve performance across the evaluated
configurations.
\vspace{-2pt}
\subsection{Effects of Visual Context on Grounding and Reasoning}
\label{sec:feedback_results}
\vspace{-2pt}
\begin{figure}[t]
  \centering
  \includegraphics[width=\linewidth]{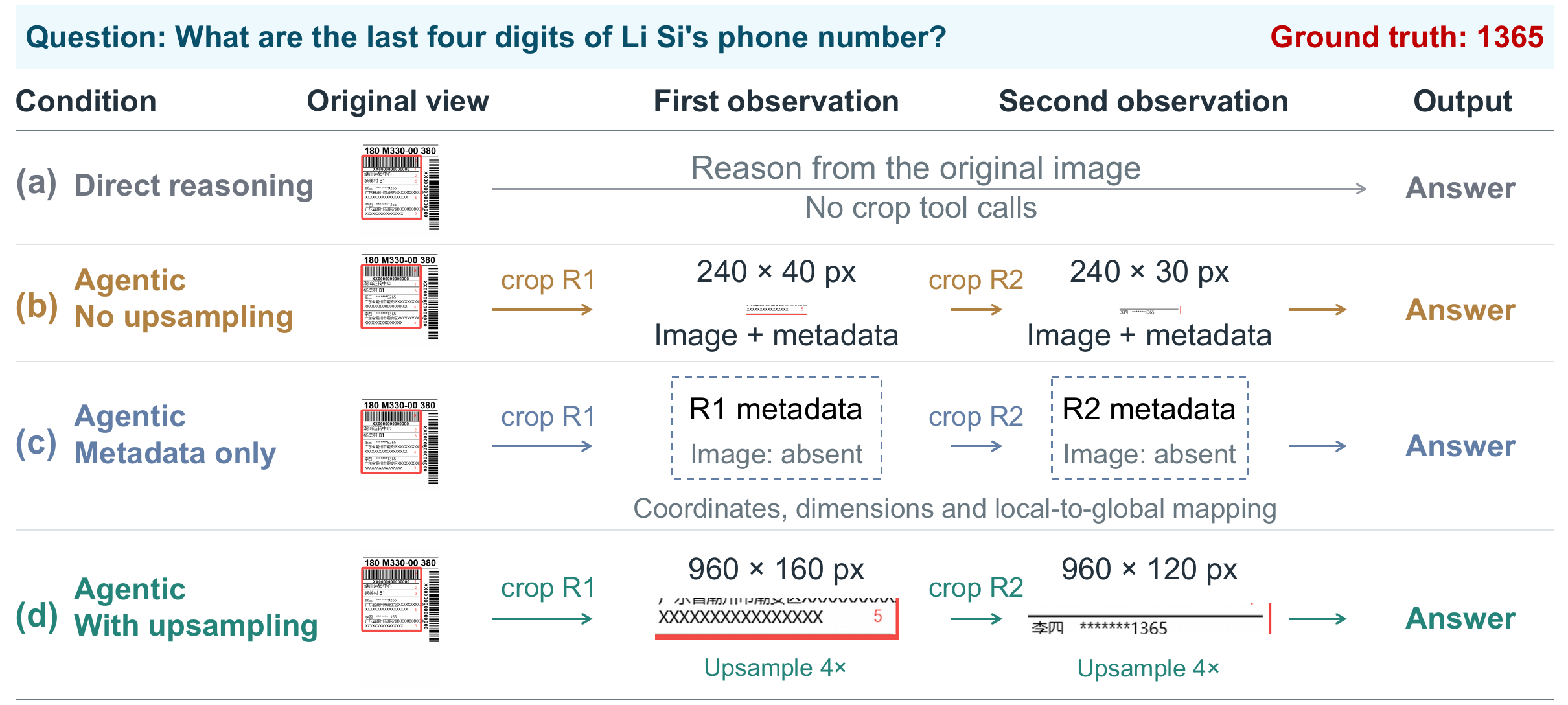}\vspace{-10pt}
\caption{Observation-feedback conditions on a text-recognition
example: original image only, native-resolution crops, crop
metadata only, and crops with $4\times$ upsampling.
All agentic contexts retain the original image and interaction
history.}
  \label{fig:observation-feedback}\vspace{-10pt}
\end{figure}
\begin{wraptable}{r}{0.55\textwidth}
\centering\vspace{-12pt}
\caption{Qwen3.8-27B answer accuracy (\%) under direct reasoning
and reconstructed observation histories.}
\label{tab:four_condition_results}
\TableSetup
\small
\setlength{\tabcolsep}{3pt}
\begin{tabularx}{\linewidth}{@{}Lcc@{}}
\toprule
Condition & \shortstack{ScreenSpot-\\Pro} & \shortstack{Perception-\\Bench} \\
\midrule
Direct reasoning (base) & 63.4 & 46.6 \\
\midrule
Crop images without upsampling & 68.8 & 52.2 \\
Crop metadata only & 70.6 & 49.9 \\
Crop images with upsampling & \textbf{81.15} & \textbf{59.67} \\
\bottomrule
\end{tabularx}
\vspace{-7pt}
\end{wraptable}

We examine how visual context affects Qwen3.8-27B answer generation
(Table~\ref{tab:four_condition_results}). From a reference interaction,
we reconstruct histories containing upsampled crops, crops without
tool-side upsampling, or crop metadata only, while retaining the
original image. These interventions modify supplied observations
without rerunning crop selection
(Appendix~\ref{sec:visual_interventions}); direct reasoning provides
a separate baseline.

The relative value of spatial metadata and crop content differs
across tasks. Metadata-only contexts outperform unupsampled crops
on ScreenSpot-Pro ($70.6\%$ vs.\ $68.8\%$), whereas crops perform
better on PerceptionBench ($52.2\%$ vs.\ $49.9\%$). This pattern
suggests that coordinate-level visual primitives support spatial
prediction in grounding, while localized crop views may better
support visual reasoning by reducing irrelevant background.
Upsampled crops achieve the highest accuracy on both benchmarks
($81.15\%$/$59.67\%$), highlighting the importance of visual detail
and presentation scale. These results characterize the use of
supplied context, rather than isolating observation behavior
acquired during RL.

\section{Conclusion}
GUI grounding provides a practical source of transferable supervision for active visual perception. Building on recent, strong open-weight VLMs and newly established baselines, we demonstrate improvements in both GUI grounding and non-GUI perception. Our results suggest that supervising intermediate observation decisions helps models inspect question-relevant regions, while crop upsampling makes fine-grained details more accessible for visual understanding. We hope this approach can be a step toward models that actively seek relevant evidence and develop a deeper understanding of the visual world.

\section{Discussion}
Our results suggest that agentic visual perception is becoming increasingly accessible to modern VLMs. With suitable supervision and a simple crop-and-inspect interface, these models can improve image-level perception without lots of additional general-domain VQA training. This progress motivates extending observation policies from individual images to video. Recent work such as Video-Index \cite{song2026video} highlights both the potential of agent tools and the remaining challenges in video understanding, while emphasizing evaluation that genuinely requires temporal evidence. A promising near-term direction is therefore to learn when and where to inspect across both space and time, including selecting informative frames and maintaining useful visual memory. Extending these capabilities to long videos and streaming inputs will further require balancing evidence acquisition with context, latency, and computation budgets.
% \clearpage
% \input{sections/statements}
\clearpage
\bibliography{main}
\clearpage
\appendix
\section{Additional Benchmark Protocols}
\label{app:breakdown}
\emph{OSWorld-G}~\cite{xie2025scaling} contains $564$ finely annotated
GUI grounding samples drawn from OSWorld rollouts, covering text
matching ($268$), element recognition ($337$), layout understanding
($252$), fine-grained manipulation ($154$), and refusal of infeasible
instructions ($54$). Unlike ScreenSpot-Pro, which uses extreme
resolutions, OSWorld-G uses standard $720$p/$1080$p screenshots and
emphasizes software commonsense and layout reasoning.

\emph{HR-Bench}~\cite{wang2024divide} evaluates perception on
high-resolution images at $4$K and $8$K resolutions, with $100$
fine-grained single-instance perception (FSP) and $100$ fine-grained
cross-instance perception (FCP) questions per resolution. The benchmark uses a cyclic-permutation protocol that averages accuracy over cyclic rotations of the answer choices.

\emph{ChartQAPro}~\cite{masry2025chartqapro} extends ChartQA with
$1{,}341$ charts from $157$ diverse sources and $1{,}948$ questions
spanning factoid, multiple-choice, conversational, fact-checking, and
hypothetical questions, including unanswerable cases. The benchmark uses an answer-type-specific score: numeric answers allow a $5\%$
relative error, years and categorical answers require exact matching,
and textual answers use average normalized Levenshtein similarity
(ANLS), following Section~4.3 and Appendix~A.5 of the benchmark paper.

\section{Implementation and Evaluation Details}
\label{app:implementation}
\subsection{Training Configuration}
\label{app:training_configuration}
Both Qwen3.5-35B-A3B and Qwen3.8-27B use the same RL optimization and rollout configuration. GRPO samples $G=8$ trajectories with temperature $0.8$ and top-$p$ $0.95$. We use AdamW~\cite{loshchilov2019decoupled} with learning rate $10^{-6}$, a cosine schedule without warmup, weight decay $0.1$, $(\beta_1,\beta_2)=(0.9,0.95)$, and gradient clipping at norm $1.0$. The global batch contains $512$ trajectories ($64$ tasks with $8$ rollouts each) across $64$ NPUs, with a per-device micro-batch size of $1$ and $8$ gradient-accumulation steps. Training uses bfloat16 precision and DeepSpeed ZeRO-3 with CPU optimizer offloading, as in the released training launcher. Token-level importance ratios use asymmetric clipping bounds $\epsilon_-=0.2$ and $\epsilon_+=0.28$. The training launcher configures one epoch, with maximum sequence length $32{,}768$, up to $12{,}288$ completion tokens per trajectory, and $2{,}560$ per assistant turn.

\subsection{Selective Masking Details}
After spatial filtering, a second filter masks intermediate turns with low mean generated-token surprisal. For each intermediate turn, this proxy is the negative mean log-probability of its generated tokens, $n_t=-\frac{1}{|\mathcal Y_t|}\sum_{k\in\mathcal Y_t}\log\pi_{\mathrm{old}}(y_{ik}\mid h_{ik})$,
where $\mathcal Y_t$ indexes the tokens of turn $t$. Among the turns that
survive the spatial filter, we mask those whose uncertainty falls below
the median uncertainty across filter-eligible turns
(equivalently, $n_t<\mathrm{median}_s\,n_s$ with an optional scaling
factor $\eta$; the main configuration uses $\eta=1$). This retains higher-uncertainty
decisions under the token-probability proxy; uncertainty does not itself
establish that a decision is useful or incorrect. The two
filters are applied sequentially over intermediate turns only; the final
assistant turn is always retained. Each intermediate mask applies
to the entire response, including reasoning and action tokens.
The spatial-masking criterion depends on terminal correctness and spatial
relevance, and the uncertainty filter depends on turn-level uncertainty;
neither depends on the sign of the group-relative advantage, so a
successful trajectory can still have a negative advantage relative to its
peers.
Masked responses remain in the conditioning history but contribute
no direct response loss. User messages and tool observations likewise
provide context without response loss. Because $Q_t$ takes a maximum,
a turn containing two crops is retained if either satisfies the
relevance threshold. Turns with only failed crops have no $Q_t$ and
are not removed by the spatial filter (the uncertainty filter still applies). Rollout log-probabilities are
aligned with the retained tokens when constructing the training payload.

\subsection{Group-Relative Optimization}\label{app:optimization}
We adapt group relative policy optimization (GRPO)~\cite{shao2024deepseekmath}
to multi-turn inspection, with the reward, token masks, and normalization
specified below. For each task, we sample $G=8$ trajectories from the rollout policy and
evaluate their rewards $R_i=R(d,\tau_i)$. Define
\begin{equation}
\bar R_d=\frac1G\sum_{i=1}^{G}R_i,
\qquad
\widehat A_i=R_i-\bar R_d.
\label{eq:advantage}
\end{equation}
The advantage subtracts the group-mean reward without variance
normalization. We disable reward scaling in Swift by setting
\texttt{scale\_rewards=none}. The aggregate trajectory reward is formed
by combining the enabled components of Equation~\eqref{eq:reward} before the
group mean is subtracted. This trajectory advantage is shared across the
retained assistant tokens.
In critic-enabled configurations, the critic enters through $R_i$,
providing an additional within-group ranking signal rather than a
separate supervised reasoning loss. Critic-enabled runs use the default settings of the critic integration in our Swift-based training pipeline, except for the activation and reward weights specified in Table~\ref{tab:ablation} and Equation~\eqref{eq:reward}. Appendix~\ref{app:prompts} describes the critic prompts and scoring procedure.
Groups with identical aggregate rewards have zero policy advantages.

Let $k$ index assistant tokens across turns, $h_{ik}$ denote the
conditioning history for token $y_{ik}$, and $M_{ik}=\widetilde m_{t(k)}$ be the combined mask of its assistant turn $t(k)$.
The current-to-old policy ratio is
\begin{equation}
r_{ik}(\theta)=
\frac{\pi_\theta(y_{ik}\mid h_{ik})}
{\pi_{\mathrm{old}}(y_{ik}\mid h_{ik})}.
\end{equation}
The policy term follows the clipped surrogate of PPO~\cite{schulman2017proximal}, with separately configured lower and upper bounds:
\begin{equation}
\ell^{\mathrm{PG}}_{ik}=
-\min\left(r_{ik}\widehat A_i,
\operatorname{clip}(r_{ik},1-\epsilon_-,1+\epsilon_+)
\widehat A_i\right),
\label{eq:clip}
\end{equation}
where $\epsilon_-$ and $\epsilon_+$ control the lower and upper
clipping bounds. Let $\pi_{\mathrm{ref}}$ be the fixed reference policy.
The reference-policy KL estimator is
\begin{equation}
\Delta_{ik}=\log\pi_{\mathrm{ref}}(y_{ik}\mid h_{ik})
-\log\pi_\theta(y_{ik}\mid h_{ik}),\qquad
\widehat K_{ik}=\exp(\Delta_{ik})-\Delta_{ik}-1,
\label{eq:kl}
\end{equation}
with the numerical safeguards of the selected trainer version.
Writing $L_i=\sum_k M_{ik}$, the sample-normalized objective is
\begin{equation}
\mathcal L_{\mathrm{RL}}=
\mathbb E_d\left[
\frac1G\sum_{i=1}^{G}
\frac{\sum_k M_{ik}
\left(\ell^{\mathrm{PG}}_{ik}+\beta\widehat K_{ik}\right)}
{\max(1,L_i)}\right].
\label{eq:rl_loss}
\end{equation}
We use Swift's \texttt{grpo} loss, which first averages over
retained tokens within each trajectory and then over trajectories.
We use the default KL coefficient $\beta$ of the Swift version used for training. KL regularization enters the
loss rather than the reward (\texttt{kl\_in\_reward=false}); hence it
does not change the rewards used in Equation~\eqref{eq:advantage}.
The token-level ratio $r_{ik}$ provides importance weighting between
the current and old policies. Under the default rollout-correction
configuration, no additional importance weight is applied to correct
rollout-engine versus trainer probability differences. Appendix~\ref{app:training_configuration} gives the clipping parameters and reported training settings.

\subsection{Crop Construction}
A requested rectangle is rounded outward to pixel boundaries and clamped
to its source image. The executor rejects degenerate crops and excessive
aspect ratios. For a crop of size $(w_c,h_c)$, the returned size is
\begin{align}
s&=\min\left(\alpha,\sqrt{\frac{P_{\max}}{w_ch_c}}\right),\\
w'&=\max(1,\lfloor sw_c\rfloor),\qquad
h'=\max(1,\lfloor sh_c\rfloor),
\label{eq:resize}
\end{align}
where $\alpha$ is the requested scale factor and $P_{\max}$ is the
pixel cap. Resizing uses Lanczos interpolation. A request on the
original image reads its source pixels; a nested request reads the
previously returned, resized crop. The executor maintains the mapping
to the original image but does not re-extract nested crops from the
original source.

\subsection{Benchmark Evaluation Protocol}
\label{sec:evaluation_protocol}

Stage-wise comparisons cover ScreenSpot-Pro and PerceptionBench for both backbones, and OSWorld-G-R, HR-Bench 4K/8K, and ChartQAPro for Qwen3.5-35B-A3B. Additional benchmark descriptions appear in Appendix~\ref{app:breakdown}.
We use benchmark-specific task definitions, reference answers, and scoring
to assess transfer from point localization to question answering.
For grounding, each example supplies a screenshot and an instruction,
and the prediction is a point in the original-image coordinate system.
Correctness is determined by inclusive point-in-box matching as in
Equation~\eqref{eq:correctness}, rather than crop IoU or exact center
matching. Invalid coordinate predictions are incorrect. The sample-level
accuracy is $N^{-1}\sum_{i=1}^{N}C(d_i,\tau_i)$, with category reporting
following the benchmark protocol.
PerceptionBench~\cite{lin2026perceptionbench} uses open-ended perception
questions with short reference answers and reference-based judgment.
We report each benchmark separately rather than combining heterogeneous scores.

Across tasks, inspection consists of alternating policy responses and
crop observations. Benchmark adapters specify the task prompt and final
answer interface: GUI tasks terminate with coordinates, whereas non-GUI
tasks terminate with the required choice or textual answer. Ground-truth
regions and training-time spatial rewards are not supplied during
evaluation. Comparisons between cold-start and RL checkpoints use the
same task interface and inference configuration within each benchmark.
Qwen3.8-27B evaluation uses reasoning effort \texttt{xhigh}, a thinking
budget of $4{,}096$ tokens per assistant turn, and at most three crop calls.

\section{Observation Feedback Protocols}
\label{sec:visual_interventions}

We compare direct reasoning with three observation conditions
constructed from an initial interaction history. All conditions
receive the same question and original input image. Direct
reasoning uses the base Qwen3.8-27B model to answer the question
without crop-tool interaction. For the other three conditions,
we first perform an initial inference run to obtain a reference
interaction history. We then manually reconstruct the historical
context to create condition-specific question--answer inputs and
evaluate the model's answers under these inputs.

The three reconstructed conditions provide crop images without
tool-side upsampling, crop metadata without image delivery, or
crop images with target scale $\alpha=4$. The original image
remains available in all conditions; the metadata-only condition
removes historical crop images without removing the original
visual input.

\paragraph{History reconstruction.}
Let $H$ denote the reference interaction history obtained from
the initial inference run. We construct three context variants:
\begin{equation}
H^{(c)}=\mathcal{T}_c(H),
\qquad
c\in\{\mathrm{no\text{-}up},\mathrm{meta},\mathrm{up}\},
\label{eq:history_intervention}
\end{equation}
where $\mathcal{T}_c$ modifies the historical crop observations
according to condition $c$. These variants share the initial
history as their source. The comparison manipulates recorded
observation content rather than evaluating three independently
generated online trajectories. Direct reasoning is evaluated
separately and is not constructed by editing $H$.

\paragraph{Crop images and spatial metadata.}
For a successful crop observation at historical turn $t$, let
$v_t^{(\alpha)}$ denote the crop image constructed with scale
parameter $\alpha$, and let $\mu_t^{(\alpha)}$ denote its
associated non-image metadata. The observation entries in the
three reconstructed contexts are
\begin{equation}
\begin{aligned}
o_t^{\mathrm{no\text{-}up}}
    &= \bigl(v_t^{(1)},\mu_t^{(1)}\bigr),\\
o_t^{\mathrm{meta}}
    &= \bigl(\varnothing,\mu_t^{(4)}\bigr),\\
o_t^{\mathrm{up}}
    &= \bigl(v_t^{(4)},\mu_t^{(4)}\bigr).
\end{aligned}
\label{eq:observation_intervention}
\end{equation}
Here, $\varnothing$ denotes the absence of a crop-image payload.
Metadata include image identifiers, dimensions, original-image
regions, resize information, and local-to-global coordinate
mappings.

The metadata-only context retains the metadata associated with
the upsampled crop observations while omitting their image
payloads. This editing rule applies to historical observations
from both original-image and nested crop requests; failure
feedback is retained. The resulting comparison examines the
value of additional crop images conditional on the original
image, spatial metadata, and the supplied interaction history.

\paragraph{Crop presentation scale.}
Following Equation~\eqref{eq:resize}, a source crop of width
$w_c$ and height $h_c$ is constructed at scale
\begin{equation}
s_{\alpha}
=
\min\left(
\alpha,\sqrt{\frac{P_{\max}}{w_c h_c}}
\right),
\qquad \alpha\in\{1,4\},
\label{eq:four_condition_scale}
\end{equation}
where $P_{\max}$ is the shared output-pixel cap. For the
no-upsampling condition,
\begin{equation}
s_{\mathrm{no\text{-}up}}
=
\min\left(
1,\sqrt{\frac{P_{\max}}{w_c h_c}}
\right).
\label{eq:no_upsampling}
\end{equation}
Crops within the pixel cap therefore retain their source
dimensions, whereas larger crops may still be downsampled.
The upsampling condition uses $\alpha=4$, enlarging each spatial
dimension by up to four times when permitted by the cap.

Both image conditions retain the model's usual image
preprocessing. Metadata describe the actual constructed views,
so image dimensions, scale factors, and coordinate mappings can
differ between the two scale settings. Interpolation changes
presentation scale without introducing new scene information.

\paragraph{Evaluation protocol.}
We evaluate the direct-reasoning baseline and the three
reconstructed-context conditions on the same evaluation items,
using the same decoding settings and applicable generation
limits. All three observation-history conditions use the same post-RL Qwen3.8-27B checkpoint and identical decoding settings and generation limits. The upsampling condition reuses the final answers and
scores from the original online evaluation of the RL model
reported in the main results, whereas answers for the
no-upsampling and metadata-only conditions are regenerated
from their modified histories. Their
scores consequently characterize answer generation under
different historical observations, rather than the outcome
of independently executed crop-selection policies.

The four evaluated conditions and their scores are reported in Table~\ref{tab:four_condition_results} in the main text.

\paragraph{Results and interpretation.}
For benchmark $b$, let $A_b^{(c)}$ denote accuracy under condition
$c$. We summarize the differences as
\begin{equation}
\begin{aligned}
\Delta_b^{\mathrm{overall}}
    &= A_b^{(\mathrm{up})}-A_b^{(\mathrm{direct})},\\
\Delta_b^{\mathrm{image}}
    &= A_b^{(\mathrm{up})}-A_b^{(\mathrm{meta})},\\
\Delta_b^{\mathrm{scale}}
    &= A_b^{(\mathrm{up})}-A_b^{(\mathrm{no\text{-}up})}.
\end{aligned}
\label{eq:four_condition_deltas}
\end{equation}
The overall contrast compares the evaluated configurations,
including differences in model checkpoint and supplied context;
it does not isolate crop-feedback effects. The image and scale
contrasts compare context variants reconstructed from the same
reference history.

Upsampled crops achieve the highest accuracy on both benchmarks,
exceeding metadata-only contexts by approximately $10.55$/$9.77$
percentage points on SSPro/PerceptionBench and unupsampled crops
by $12.35$/$7.47$ points. These results highlight the importance
of both crop-image availability and presentation scale.

The comparison between metadata-only and unupsampled-crop
contexts reveals a task-dependent pattern. Metadata-only
contexts perform better on SSPro ($70.6\%$ versus $68.8\%$),
whereas unupsampled crops perform better on PerceptionBench
($52.2\%$ versus $49.9\%$). This pattern suggests that
coordinate-level visual primitives can support spatial
prediction in grounding, while localized image content may
aid non-GUI reasoning by reducing irrelevant background.
The strongest results with upsampled crops suggest
that both tasks benefit from suitably presented local visual
evidence.

\section{Extended PerceptionBench Reference Results}
Table~\ref{tab:pb_external} reports external reference results on PerceptionBench.
\label{app:pb_external}
\vspace{-7pt}
\begin{table}
\centering
\caption{PerceptionBench accuracy (\%) across ten perceptual capabilities.
Panels (a,b) provide results for 16 external reference models.}
\label{tab:pb_external}
\TableSetup
\setlength{\tabcolsep}{2pt}
\renewcommand{\arraystretch}{1.0}
\noindent\textbf{(a) External reference models (1--8)}\par\smallskip
\begin{tabularx}{\linewidth}{@{}L *{8}{N{36.5pt}}@{}}
\toprule
Capability & {\parbox[t]{36.5pt}{\centering GPT-5.6-\\Sol\strut}} & {\parbox[t]{36.5pt}{\centering Kimi\\K3\strut}} & {\parbox[t]{36.5pt}{\centering Claude-\\Fable-5\strut}} & {\parbox[t]{36.5pt}{\centering Gemini-\\3.1-Pro\strut}} & {\parbox[t]{36.5pt}{\centering GPT-5.5\strut}} & {\parbox[t]{36.5pt}{\centering Seed-\\2.1-Pro\strut}} & {\parbox[t]{36.5pt}{\centering Gemini-\\3.5-Flash\strut}} & {\parbox[t]{36.5pt}{\centering Qwen\\3.7-Plus\strut}} \\
\midrule
VRel & \bfseries 69.7 & 68.2 & 58.5 & 58.8 & 61.9 & 57.6 & 53.6 & 59.1 \\
Count & \bfseries 62.4 & 59.7 & 52.9 & 56.9 & 55.8 & 51.2 & 43.6 & 53.3 \\
Attr & \bfseries 62.1 & 59.4 & 60.9 & 61.8 & 60.9 & 58.2 & 54.5 & 55.8 \\
Depth & \bfseries 55.5 & 52.4 & 51.5 & 50.0 & 48.8 & 43.6 & 49.7 & 48.5 \\
Loc & \bfseries 76.7 & 70.3 & 70.4 & 52.7 & 65.8 & 50.0 & 50.6 & 52.7 \\
Comp & \bfseries 67.0 & 59.1 & 56.1 & 61.7 & 65.6 & 59.5 & 54.8 & 55.9 \\
FGR & 55.9 & 55.9 & 51.6 & 54.8 & 47.2 & \bfseries 56.6 & 51.7 & 46.8 \\
Ctx & 60.0 & 53.3 & 59.8 & \bfseries 61.2 & 58.0 & 60.4 & 53.3 & 52.2 \\
OCR & 54.9 & 61.2 & 64.3 & 64.3 & 56.5 & \bfseries 66.7 & 59.6 & 54.5 \\
Hallu & 26.9 & 41.7 & 45.0 & 40.6 & 34.7 & 49.8 & \bfseries 50.6 & 29.5 \\
\midrule
\textbf{Overall} & \bfseries 59.7 & 58.5 & 57.2 & 56.2 & 55.8 & 55.0 & 52.0 & 51.1 \\
\bottomrule
\end{tabularx}
\par\vspace{6pt}
\noindent\textbf{(b) External reference models (9--16)}\par\smallskip
\begin{tabularx}{\linewidth}{@{}L *{8}{N{36.5pt}}@{}}
\toprule
Capability & {\parbox[t]{36.5pt}{\centering Qwen\\3.5-397B-\\A17B\strut}} & {\parbox[t]{36.5pt}{\centering Claude-\\Opus-4.8\strut}} & {\parbox[t]{36.5pt}{\centering Kimi\\K2.6\strut}} & {\parbox[t]{36.5pt}{\centering Grok-4.5\strut}} & {\parbox[t]{36.5pt}{\centering Gemma-\\4-31B\strut}} & {\parbox[t]{36.5pt}{\centering GLM-5V\\-Turbo\strut}} & {\parbox[t]{36.5pt}{\centering Minimax\\-M3\strut}} & {\parbox[t]{36.5pt}{\centering GLM-\\4.6V\strut}} \\
\midrule
VRel & 55.2 & 51.4 & 50.9 & 47.0 & 42.7 & 41.2 & 40.0 & 35.2 \\
Count & 49.1 & 44.2 & 45.2 & 35.2 & 33.9 & 40.0 & 30.3 & 31.8 \\
Attr & 53.0 & 49.4 & 43.6 & 39.4 & 40.3 & 41.2 & 34.6 & 35.2 \\
Depth & 44.6 & 40.6 & 42.4 & 41.2 & 39.1 & 41.2 & 36.7 & 29.1 \\
Loc & 46.7 & 58.8 & 45.2 & 39.7 & 44.9 & 43.9 & 33.3 & 30.6 \\
Comp & 49.8 & 48.4 & 39.1 & 43.7 & 43.7 & 45.2 & 31.2 & 34.8 \\
FGR & 44.8 & 40.7 & 34.5 & 36.2 & 39.0 & 36.6 & 26.6 & 29.3 \\
Ctx & 50.2 & 44.7 & 40.4 & 39.6 & 45.9 & 32.9 & 31.0 & 33.7 \\
OCR & 52.9 & 54.1 & 40.8 & 43.9 & 46.7 & 43.5 & 35.7 & 39.2 \\
Hallu & 26.9 & 38.7 & 41.0 & 44.7 & 32.1 & 28.0 & 29.9 & 26.9 \\
\midrule
\textbf{Overall} & 47.5 & 47.2 & 42.6 & 41.0 & 40.7 & 39.6 & 33.1 & 32.5 \\
\bottomrule
\end{tabularx}
\par\vspace{6pt}
\TableNote{External scores: PerceptionBench~\cite{lin2026perceptionbench},
\href{https://github.com/MoonshotAI/PerceptionBench}{official leaderboard} (accessed September 21, 2026).
Evaluation and inference settings may differ.
Bold in (a,b) marks the highest external score per capability across all 16 models.
Our checkpoint results appear in Table~\ref{tab:pb_results_combined}.}
\TableNote{VRel: visual relation; Count: counting; Attr: attribute; Depth: depth/3D perception;
Loc: localization; Comp: comparison; FGR: fine-grained recognition; Ctx: contextual integration;
OCR: optical character recognition; Hallu: perception-related hallucination.}
\end{table}

\clearpage
\section{Example of a complete agentic grounding/reasoning trace}
\begin{figure}[H]
  \centering
  \includegraphics[width=0.95\linewidth,height=0.76\textheight,keepaspectratio]{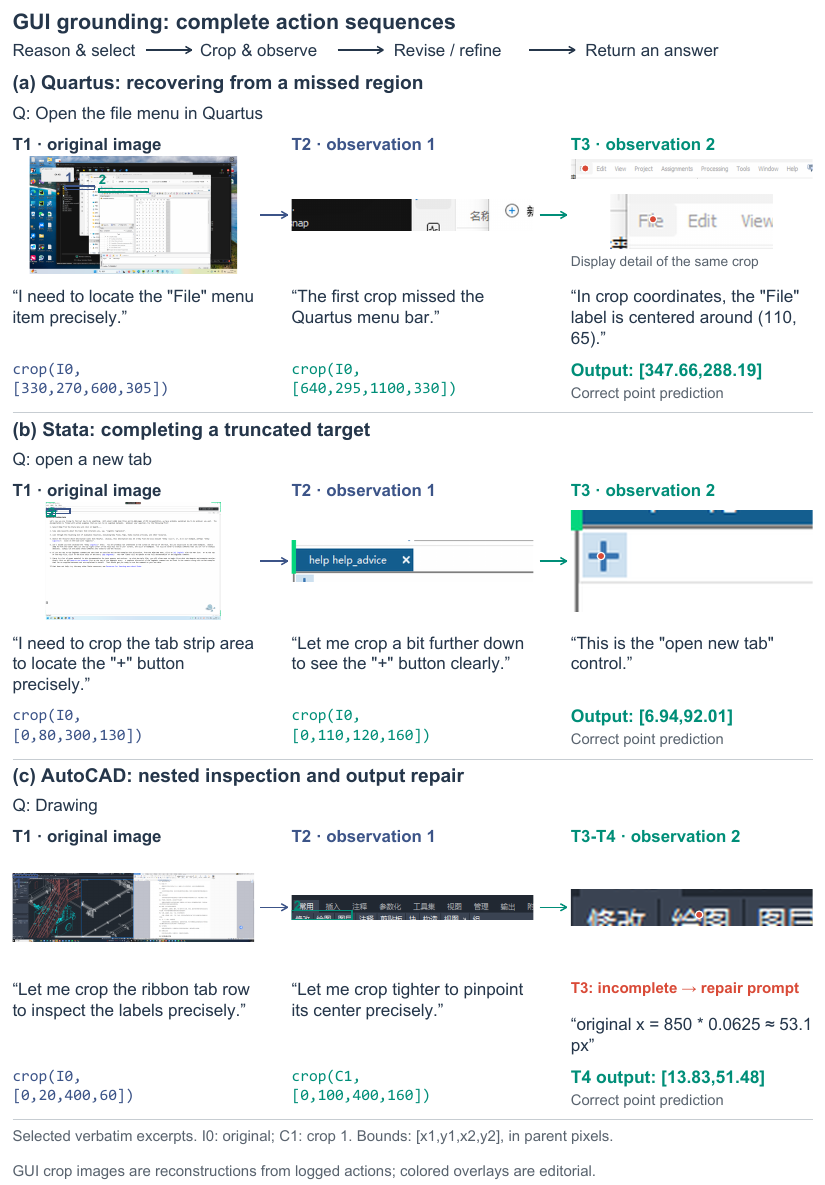}
  \caption{Recorded GUI observation trajectories with verbatim reasoning excerpts and all crop actions. Quartus relocates a missed region, Stata refines a truncated target, and AutoCAD recursively crops a returned view and repairs an incomplete response. I0 denotes the original image and C1 the first crop. GUI observations are reconstructed from logged actions; colored annotations are editorial.}
  \label{fig:gui-observation-traces}
\end{figure}
\begin{figure}[p]
  \centering
  \includegraphics[width=0.95\linewidth,height=0.82\textheight,keepaspectratio]{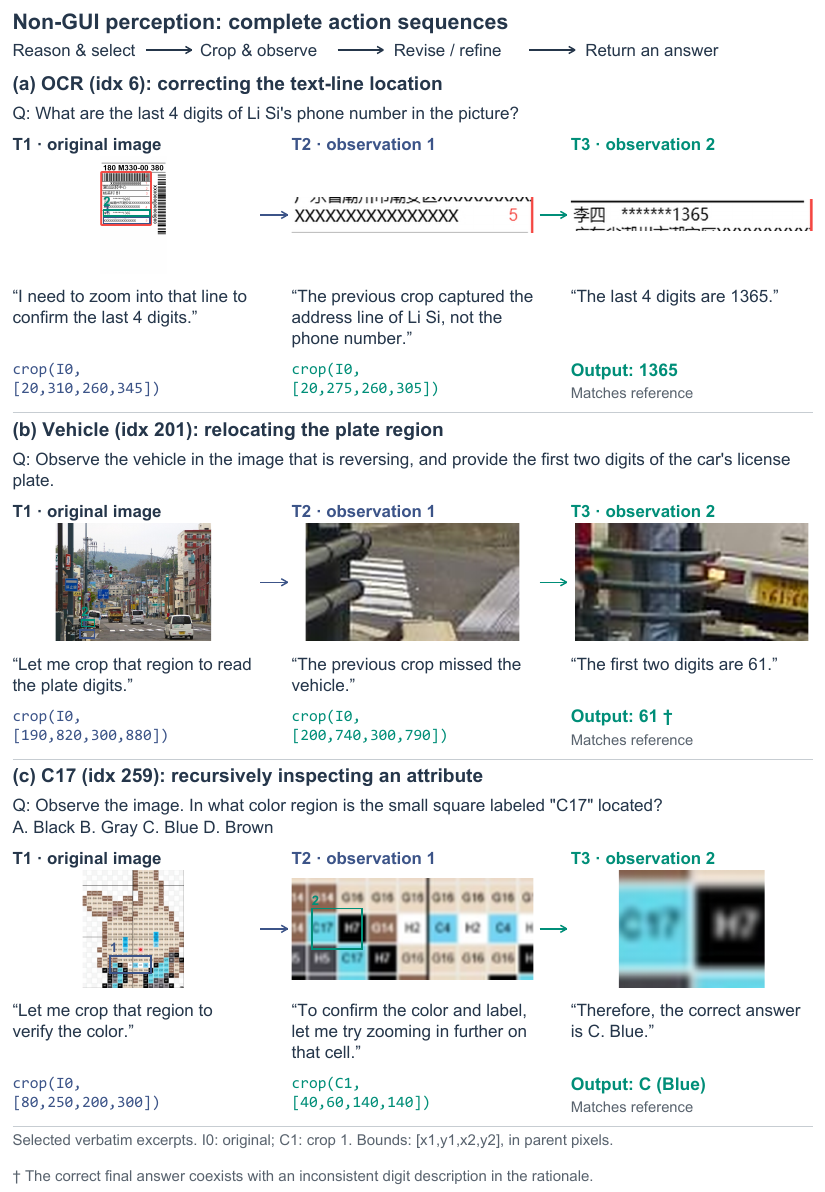}
  \caption{Complete PerceptionBench action sequences with verbatim reasoning excerpts. OCR and vehicle recognition revise misplaced crops; C17 recursively inspects a returned observation. Bounds use parent-image pixels. All final answers match their references.}
  \label{fig:pb-observation-traces}
\end{figure}
\clearpage
\section{Training and Critic Prompts}
\label{app:prompts}
\tcbset{paperprompt/.style={
  breakable,
  colback=black!2,colframe=black!55,
  colbacktitle=black!8,coltitle=black,
  fonttitle=\bfseries,boxrule=0.5pt,arc=1mm,
  left=6pt,right=6pt,top=5pt,bottom=5pt,
  before skip=10pt,after skip=10pt
}}

\subsection{System Prompt for Agentic RL}
\label{sec:p_agentic_rl}

We train with ms-swift GRPO and a custom multi-turn scheduler.
At rollout initialization, the scheduler constructs the system message
below from the protocol template, expanding the crop description.
Braced fields are filled with the task, image metadata, and tool budget;
line wrapping is for presentation only. The original screenshot is
supplied in the initial user message.

The \texttt{\{max\_tool\_calls\}} field is set to the configured
tool-call budget, capped at $3$ (default: $3$).
The scheduler executes textual crop actions and returns images and metadata
in a user message, without a native function-calling schema. GRPO rewards
and loss masks are computed by the framework and plugin, not specified
through an additional training prompt.

\begin{tcolorbox}[paperprompt,title={Agentic RL system prompt $p_{\mathrm{RL}}$},title after break={Agentic RL system prompt $p_{\mathrm{RL}}$ (continued)}]
\begin{Verbatim}[fontsize=\small,breaklines=true,breakanywhere=true,breaksymbolleft={},breaksymbolright={}]
You are an agentic GUI grounding model.

Task instruction:
{instruction}

Original image metadata:
- image_path: {image_path}
- resolution: {width}x{height} pixels (width x height)
- origin: [0,0] at the top-left

Goal:
Output the center point of the position corresponding to the task
instruction. The final coordinate MUST be relative to the ORIGINAL
image and normalized to [0.00,1000.00], where [0.00,0.00] is the
original image's top-left and [1000.00,1000.00] is its bottom-right.
Never output absolute pixels or coordinates local to a crop.
Preserve sub-unit precision:
the two final coordinate values MUST each have exactly two digits
after the decimal point (for example, [512.34,76.50]). If the task
is infeasible (for example, the requested UI element or action is
absent from the image), use [-1.00,-1.00].

Available local action:
- crop_image: Crop an image that has already been supplied in this
  conversation. image_path must be the exact original path or a path
  returned by an earlier crop_image call. x1, y1, x2, y2 are
  absolute pixel coordinates in that referenced image's LOCAL
  coordinate system, with (x1,y1) at the top-left and (x2,y2) at the
  bottom-right. The result provides a new image path, dimensions,
  and the crop region mapped into the ORIGINAL image. Returned crops
  are resized for inspection, so nested crops must use the returned
  width and height. Crops may be nested. The final answer must
  always use relative coordinates of the ORIGINAL image normalized
  to the [0,1000] range, never crop-local coordinates. When writing
  a crop_image action, format each of x1, y1, x2, and y2 with
  exactly two digits after the decimal point.

Text-action protocol:
Every turn must first emit visible reasoning inside
<think>...</think>, then end with
exactly one of these choices:
- To inspect one region:
  <think>brief reasoning for the crop</think>
  <action>crop_image(image_path="exact path", x1=100.00, y1=200.00, x2=500.00, y2=600.00)</action>
- To inspect two independent regions in the same turn:
  <think>brief reasoning for both crops</think>
  <action>crop_image(image_path="exact path", x1=100.00, y1=200.00, x2=500.00, y2=600.00)</action>
  <action>crop_image(image_path="exact path", x1=600.00, y1=200.00, x2=900.00, y2=600.00)</action>
- To return the final point:
  <think>brief reasoning for the final point</think>
  [x.xx,y.yy]

crop_image is a textual action interpreted locally after your
answer. It is NOT an API function call. Do not emit JSON,
<tool_call> XML, or native tool_calls. Keep x1, y1, x2, and y2 as
four separately named numeric arguments.

Tool rules:
1. Before returning the final coordinate, you SHOULD make at least
   one successful crop_image call to inspect the target more
   precisely. A direct first-turn coordinate is discouraged. Skip
   cropping only when the task is clearly infeasible or cropping
   cannot reasonably improve localization.
2. You may make at most {max_tool_calls} crop_image calls in total.
   Never exceed this budget; use each crop purposefully and return
   the final coordinate as soon as the target is localized. A single
   assistant turn may contain one or two crop_image actions, and
   every action counts separately toward the total budget. Never
   emit more than two actions in one turn or mix crop actions with a
   final coordinate.
3. Tool arguments are local absolute pixels in the image_path passed
   to that call. Write x1, y1, x2, and y2 with exactly two digits
   after the decimal point, including trailing zeros (for example,
   x1=100.00). Two actions emitted in the same turn must refer only
   to image paths already available before that turn; the second
   action cannot depend on the result of the first action from the
   same turn.
4. Each result states the crop's region both in original-image
   pixels and original-image normalized [0,1000] coordinates. Use
   that mapping to convert any point found in a crop back to the
   ORIGINAL image's normalized [0,1000] coordinate system.
5. After crop_image actions, the following single user message
   contains all result metadata and all successful crop images in
   action order.

Final-answer rule:
When ready, output exactly <think>...</think> followed by a newline
and [x.xx,y.yy], with no prose, Markdown, or action wrapper after
the coordinate. Both values must contain exactly two digits after
the decimal point, including trailing zeros. The point must be in
the ORIGINAL image's normalized [0,1000] coordinate system. If the
task is infeasible, use [-1.00,-1.00].
\end{Verbatim}
\end{tcolorbox}

\paragraph{Initial image message and protocol correction.}
The initial user message uses ms-swift's \texttt{<image>} marker, with
the screenshot supplied through the request's image field. Unparseable
responses trigger the correction below as a user turn, with a default
allowance of $2$ corrections. Crop observations follow the system protocol.

\begin{tcolorbox}[paperprompt,title={Initial user message $p_{\mathrm{image}}$},title after break={Initial user message $p_{\mathrm{image}}$ (continued)}]
\begin{Verbatim}[fontsize=\small,breaklines=true,breakanywhere=true,breaksymbolleft={},breaksymbolright={}]
image_path: {image_path}<image>
\end{Verbatim}
\end{tcolorbox}

\begin{tcolorbox}[paperprompt,title={Invalid-action correction $p_{\mathrm{correct}}$},title after break={Invalid-action correction $p_{\mathrm{correct}}$ (continued)}]
\begin{Verbatim}[fontsize=\small,breaklines=true,breakanywhere=true,breaksymbolleft={},breaksymbolright={}]
Your previous answer was not valid. Reply with <think>...</think>
followed by either one or two independent crop actions, each in its
own <action>crop_image(...)</action> block, or one final [x.xx,y.yy]
coordinate. Do not emit more than two actions or mix actions with a
final coordinate. Every coordinate value must have exactly two
digits after the decimal point.
\end{Verbatim}
\end{tcolorbox}

\paragraph{Prompting and learned behavior.}
The system prompt explicitly encourages at least one successful crop and
requests a final answer once the target is localized. These instructions
are part of the interaction protocol; adherence alone is not evidence
that RL learned when inspection is useful or when to stop.

\subsection{Prompts for CoT Scoring}
\label{sec:p_cot_critic}

The critic receives a separate text-only request. In blind mode,
it sees the task and transcript,
with the five criteria below substituted for \texttt{\{criteria\}}, but
no images or target annotations. It scores textual reasoning--action
consistency and cannot verify visual claims. The boxes use the supplied
English rendering; the Chinese implementation requests an $80$-character
comment, whereas this rendering specifies $80$ words.

\begin{tcolorbox}[paperprompt,title={DeepSeek critic system prompt $p_{\mathrm{cot}}^{\mathrm{sys}}$ (blind mode)},title after break={DeepSeek critic system prompt (continued)}]
\begin{Verbatim}[fontsize=\small,breaklines=true,breakanywhere=true,breaksymbolleft={},breaksymbolright={}]
You are a quality reviewer of the reasoning process of a GUI
grounding agent. The agent's task is to locate the center point of
the target described by a task instruction within a UI screenshot;
it may first crop and zoom into the image multiple times to inspect
it, and must finally output coordinates normalized to [0, 1000], or
[-1.00, -1.00] when the task is infeasible.

You can only see text: the task instruction, the agent's per-turn
thinking (think), action decisions (action), and the execution
results of crop actions (feedback). You cannot see the images, and
you do not need to solve the task yourself -- only evaluate the
quality of the reasoning process.

Evaluation criteria:
1. Goal orientation: whether each turn's thinking revolves around
   "how to locate the target of the task instruction";
2. Feedback utilization: whether subsequent thinking leverages the
   results of previous crop executions (successful region ranges /
   failure reasons);
3. Thought-action consistency: whether the regions and directions
   described in the thinking match the coordinate parameters in the
   actions;
4. Incremental progress: whether each turn brings new information or
   new decisions, without repeating the same crop or reasoning;
5. Conciseness: no meaningless filler or parroting.

Scoring anchors:
- 0.9~1.0: The thinking stays focused on locating the target
  throughout, fully leverages the crop feedback from each turn, is
  consistent with the action parameters, converges step by step, and
  is concise without redundancy;
- 0.6~0.8: The thinking is generally reasonable, with only minor
  flaws that do not affect the main line (e.g., slightly imprecise
  coordinate descriptions);
- 0.3~0.5: There are obvious problems -- at least one of: ignoring
  tool feedback, inconsistency between thinking and actions,
  repeating the same ineffective operation, or reasoning detached
  from the task goal;
- 0.0~0.2: Almost no effective reasoning, heavy verbatim repetition,
  or content completely irrelevant to the task.

First give a brief comment of no more than 80 words, then output the
score alone on a new
line, in exactly this format:
Score: 0.xx
(a float in 0.0~1.0, rounded to two decimal places)
\end{Verbatim}
\end{tcolorbox}

\begin{tcolorbox}[paperprompt,title={Critic user prompt $p_{\mathrm{cot}}^{\mathrm{user}}$},title after break={Critic user prompt $p_{\mathrm{cot}}^{\mathrm{user}}$ (continued)}]
\begin{Verbatim}[fontsize=\small,breaklines=true,breakanywhere=true,breaksymbolleft={},breaksymbolright={}]
Task instruction: {task}

The agent's full trajectory (think = reasoning, action = action
decision, feedback = crop
execution result):
{transcript}
\end{Verbatim}
\end{tcolorbox}

\paragraph{Transcript construction and reward use.}
Each serialized assistant turn contains \texttt{[Turn N]}, \texttt{think:},
\texttt{action:}, and available \texttt{feedback:} fields. Feedback records
crop geometry and dimensions or execution failures; images are excluded.
Default character limits are $1{,}500$ per reasoning segment, $600$ per
action summary, and $6{,}000$ per transcript, so ``full trajectory'' is
subject to truncation. The default reward multiplies the parsed score by
terminal correctness (Equation~\eqref{eq:critic}), without exposing that
label to the blind critic. Trajectories without visible reasoning receive zero.

\paragraph{Optional oracle variant.}
When oracle mode is enabled, the sixth criterion and
reference annotation below are appended to the system and user prompts,
respectively. Other instructions are unchanged. The annotation is exposed
to the critic, not the policy; the critic still receives no images.

\begin{tcolorbox}[paperprompt,title={Oracle addition to the critic system prompt},title after break={Oracle addition to the critic system prompt (continued)}]
\begin{Verbatim}[fontsize=\small,breaklines=true,breakanywhere=true,breaksymbolleft={},breaksymbolright={}]
6. Annotation alignment: against the reference annotation (GT bbox
   or infeasibility flag), whether the cropped regions and
   coordinate directions are progressively approaching the correct
   location; for refusal samples, evaluate whether the justification
   for judging the task as infeasible is sufficient.
\end{Verbatim}
\end{tcolorbox}

\begin{tcolorbox}[paperprompt,title={Oracle addition to the critic user prompt},title after break={Oracle addition to the critic user prompt (continued)}]
\begin{Verbatim}[fontsize=\small,breaklines=true,breakanywhere=true,breaksymbolleft={},breaksymbolright={}]
Reference annotation (GT bbox in [0, 1000] normalized coordinates;
[-1, -1, -1, -1] means the task is infeasible): {ground_truth}
\end{Verbatim}
\end{tcolorbox}

\end{document}